\documentclass[letterpaper]{article} 
\usepackage[submission]{aaai23}  
\usepackage{times}  
\usepackage{helvet}  
\usepackage{courier}  
\usepackage[hyphens]{url}  
\usepackage{graphicx} 
\usepackage{natbib}  
\usepackage{caption} 
\usepackage[linesnumbered,ruled,vlined]{algorithm2e}

\usepackage{float}
\usepackage{listings}
\DeclareCaptionStyle{ruled}{labelfont=normalfont,labelsep=colon,strut=off} 
\floatstyle{ruled}
\newfloat{listing}{tb}{lst}{}
\floatname{listing}{Listing}

\usepackage[utf8]{inputenc} 
\usepackage{booktabs}       
\usepackage{amsfonts}       
\usepackage{nicefrac}       
\usepackage{microtype}      
\usepackage{xcolor}         
\usepackage{subcaption}
\usepackage{amsmath}
\usepackage{amsthm}
\usepackage{enumerate}
\usepackage{bm}
\usepackage{bbm}

\title{Soft Action Priors: Towards Robust Policy Transfer}
\author {
    Matheus Centa\textsuperscript{\rm 1}
    Philippe Preux, \textsuperscript{\rm 1}
}
\affiliations {
    \textsuperscript{\rm 1} Univ. Lille, CNRS, Inria\\
    \{matheus.medeiros-centa, philippe.preux\}@inria.fr
}

\begin{document}

\maketitle

\begin{abstract}
Despite success in many challenging problems, reinforcement learning (RL) is still confronted with sample inefficiency, which can be mitigated by introducing prior knowledge to agents. However, many transfer techniques in reinforcement learning make the limiting assumption that the teacher is an expert. In this paper, we use the action prior from the Reinforcement Learning as Inference framework \cite{levine2018reinforcement} - that is, a distribution over actions at each state which resembles a teacher policy, rather than a Bayesian prior - to recover state-of-the-art policy distillation techniques. Then, we propose a class of adaptive methods that can robustly exploit action priors by combining reward shaping and auxiliary regularization losses. In contrast to prior work, we develop algorithms for leveraging suboptimal action priors that may nevertheless impart valuable knowledge - which we call \emph{soft action priors}. The proposed algorithms adapt by adjusting the strength of teacher feedback according to an estimate of the teacher's usefulness in each state. We perform tabular experiments, which show that the proposed methods achieve state-of-the-art performance, surpassing it when learning from suboptimal priors. Finally, we demonstrate the robustness of the adaptive algorithms in continuous action deep RL problems, in which adaptive algorithms considerably improved stability when compared to existing policy distillation methods.
\end{abstract}

\section{Introduction}

Model-free reinforcement learning has been successfully employed in a wide range of problem domains \cite{silver2014deterministic,schulmanetal2016high,lillicrap2015continuous,mnih2016asynchronous}. However, these methods demand a significant amount of experience to achieve state-of-the-art results which effectively limits their usefulness in some domains \cite{thompson2020computational}. One such domain is healthcare, in which one of the biggest challenges is learning with few samples \cite{yu2021reinforcement,gottesman2019guidelines}. Moreover, in many applications, researchers have access to \emph{some} knowledge that could be used to speed-up learning. 
By that, we mean that this information does not necessarily convey optimal or ``expert'' knowledge. Our goal is to leverage such information to improve the sample efficiency of RL algorithms in a principled and robust way. Specifically, we study leveraging a \textit{soft action prior} $\pi_0$ which is a policy-like function mapping states to distributions over actions. While other common transfer methods for Deep RL assume that the transferred knowledge is given in the form of trajectories \cite{hester2018deep, nair2018overcoming} or the underlying function approximators \cite{devin2017learning}, we assume that the transferred knowledge is the soft action prior $\pi_0$. 

In this paper, we revisit the Reinforcement Learning as Inference \cite{levine2018reinforcement} framework, in which \emph{action priors} $\pi_0 (.|\bm{s})$ appear naturally in the proposed probabilistic model. When using this framework, it is common to assume non-informative action priors, that is, uniform probability distributions over the action space. We show how this assumption can be alleviated by reintroducing non-uniform action priors to RL algorithms.

Specifically, we introduce a \emph{prior trick} to be used to obtain a reward bonus that recovers the effects of non-uniform priors. The resulting bonus coincides with the family of \emph{entropy regularized} policy distillation algorithms \cite{distral,SchulmanAC17,czarnecki2019distilling}. Next, we design a class of adaptive policy distillation algorithms for the soft action prior setting. Our main contribution is weighing rewards bonuses with \emph{prior weights} which depend on the state. Such weights are learned parameters and control how much the prior shapes the reward at each state.

Despite similarities with other policy distillation methods, the proposed adaptive algorithm is able to better leverage imperfect teachers. As a result, the proposed methods are more widely applicable to scenarios where acquiring expert knowledge is costly, but some helpful knowledge is available. Our central assumption is that the prior may not be optimal, but it is nevertheless informative about good actions in some states. Consequently, following these priors may not result in trajectories with high returns, but they are helpful when learning optimal policies.

We propose a tabular experimental setup to study how well different algorithms exploit different types of noisy priors, which we call \textit{degraded priors}. Empirically, we find that adaptive algorithms match state-of-the-art performance in the expert setting and surpass it when learning from imperfect priors. We also find that the performance gain is larger in more degraded settings.

Finally, an empirical study on continuous control benchmarks using deep actor-critics validates the advantages of adaptive methods in the deep RL setting. Our proposed algorithm significantly improve upon the robustness of existing methods while also outperforming them in the majority of the considered environments. Our results show that robust policy distillation is advantageous when leveraging priors in complex tasks.

Our contributions can be summarized as follows:
\begin{enumerate}[(i)]
    \item First, we show that a prior trick can be used to derive state-of-the-art policy distillation methods.
    \item Second, we propose a class of adaptive algorithms that can adjust the action prior's relevancy as a function of the state.
    \item Then, we show that the proposed methods match or surpass previous state-of-the-art performances in several tabular experiments, significantly improving transfer performance when using suboptimal priors.
    \item Finally, we empirically compare performance in continuous control benchmarks to show that adaptive methods are significantly more robust than existing methods in realistic settings.
\end{enumerate}

\section{Preliminaries}

We consider a finite-horizon Markov Decision Problem \cite{puterman94}, which is defined as the tuple $\mathcal{M} = \langle \mathcal{S}, \mathcal{A}, R, P, T \rangle$, with horizon $T$, state space $\mathcal{S}$, action space $\mathcal{A}$, bounded reward function $R$, dynamics $P$. The agent interacts with the environment at time $t$ by observing state $\bm{s}_t \in \mathcal{S}$ and taking an action $\bm{a}_t \sim \pi(. |\bm{s}_t), \ \bm{a}_t \in \mathcal{A}$, after which it observes a reward $r_t \in \mathbb{R}$. The next state $\bm{s}_{t+1}$ is sampled according to $P(\bm{s}_t, \bm{a}_t, \bm{s}_{t+1})$. We ignore discount factors in the derivation of the reward bonus, but it can be recovered adding an absorbing state and modifying the dynamics \cite{levine2018reinforcement}.

Our goal is to find a policy $\pi$ that optimizes $J(\pi) = \mathbb{E}_{\pi, \mathcal{M}} [\sum_{t=0}^T R(\bm{s}_t, \bm{a}_t)]$. The state value $V^\pi(s) = \mathbb{E}_{\pi, \mathcal{M}} [ \sum_{t=0}^T R(\bm{s}_t, \bm{a}_t) | \bm{s}_0 = s]$ is the expected cumulative reward achieved by $\pi$ from state $s$. Similarly, the state-action value $Q^\pi(s, a) = \mathbb{E}_{\pi, \mathcal{M}} [ \sum_{t=0}^T R(\bm{s}_t, \bm{a}_t) | s_0 = \bm{s}, \bm{a}_0 = a ]$ denotes the expected cumulative reward achieved by $\pi$ after taking action $a$ at state $s$. The entropy of a policy at state $s$ is the quantity $H^\pi(s) = \mathbb{E}_{\pi(.|s)} [ - \log \pi(.|s) ]$ and the KL divergence and cross-entropy between two policies at state $s$ are the quantities $D_{s}^{KL} (\pi || \pi') = \mathbb{E}_{\pi(.|s)} \left[ \log \frac{\pi(.|s)}{\pi'(.|s)} \right]$ and $H_{s}^{X} (\pi || \pi') = - \mathbb{E}_{\pi(.|s)} \left[ \log \pi'(.|s) \right]$, respectively.

\subsection{Reinforcement Learning as Inference}

We take inspiration from the Reinforcement Learning as Inference Framework, reviewed in \cite{levine2018reinforcement}, which reframes the computation of the optimal policy as an inference problem. In order to achieve this, \emph{optimality variables} $\mathcal{O}_t$ are introduced: in their simplest formulation, these are Bernoulli variables such that they are 1 (or \texttt{True}) when the action $\bm{a}_t$ taken at state $\bm{s}_t$ is \emph{optimal}. In most formulations, this notion optimality translates into the assumption that $\mathbb{P}(\mathcal{O}_t | \bm{s}_t, \bm{a}_t) \propto \exp R(\bm{s}_t, \bm{a}_t)$. Then, computing the optimal policy can be reframed as inference of $\mathbb{P}(\bm{a}_t | \bm{s}_t, \mathcal{O}_{1:T})$, where $\mathcal{O}_{1:T}$ indicates that the optimality variables are \texttt{True} for the entire trajectory. Note that this notion of optimality is \emph{local} and, as a result, we seek policies that maximize the probability of $\mathcal{O}_{1:T}$ (as opposed to $\mathcal{O}_t$).

In this formulation, when solving the inference problem 
with a standard sum-product inference algorithm, we compute backward messages of the form:
\begin{align*}
    \beta_t^{\pi_0}(\bm{s}_t, \bm{a}_t) &:= \mathbb{P}(\mathcal{O}_{t:T}|\bm{s}_t, \bm{a}_t), \\
    \beta_t^{\pi_0}(\bm{s}_t) &:= \mathbb{P}(\mathcal{O}_{t:T}|\bm{s}_t) \\ &= \mathbb{E}_{\bm{a}_t \sim \pi_0(.|\bm{s}_t)} \left[\beta_t^{\pi_0}(\bm{s}_t, \bm{a}_t)\right],
\end{align*}

which resemble state-action and state value functions in logspace respectively. Note that during the inference procedure, a choice of $\mathbb{P}(\bm{a}_t | \bm{s}_t)$ must be made. This is not necessarily the optimal policy $\pi^* := \mathbb{P}(\bm{a}_t | \bm{s}_t, \mathcal{O}_{1:T})$, but rather acts as a prior over actions. We refer to such probabilities as the action prior $\pi_0$.

\subsection{Prior Trick}

Authors usually ignore this action prior in their derivations by setting it to the uniform distribution \cite{levine2018reinforcement}. The mathematical trick commonly used to justify this simplification is the \emph{prior trick}, which simply performs importance weighing to return to the uniform case: $\beta_t^{\pi_0}(\bm{s}_t) = \mathbb{E}_{\bm{a}_t \sim \mathcal{U}_\mathcal{A}} \left[\pi_0(\bm{a}_t | \bm{s}_t) \beta_t^{\pi_0}(\bm{s}_t, \bm{a}_t)\right]$. However, these backward messages have a neat interpretation as value functions in logspace. Combining the definitions above, we easily get:
\begin{gather}
    V^{\pi_0}(\bm{s}_t) = \mathbb{E}_{\bm{a}_t \sim \mathcal{U}_\mathcal{A}} \left[\log \pi_0(\bm{a}_t | \bm{s}_t)+ \log \beta_t^{\pi_0}(\bm{s}_t, \bm{a}_t)\right] \nonumber\\
    V^{\pi_0}(\bm{s}_t) = \mathbb{E}_{\bm{a}_t \sim \mathcal{U}_\mathcal{A}} \left[\log \pi_0(\bm{a}_t | \bm{s}_t)+ Q^\pi_0 (\bm{a}_t | \bm{s}_t) \right] \nonumber\\
\end{gather}

where $\mathcal{U}_\mathcal{A}$ is the uniform distribution over the action space. As a result, one can absorb the term $\log \pi_0 (\bm{a}_t | \bm{s}_T)$ into the Q-function or, more generally, the reward.

\section{Recovering the Action Prior}

We aim to introduce the action prior into RL algorithms. The simplest way to achieve this is by adding the reward bonus $B^{ER}(\bm{s}_t, \bm{a}_t) := \log \pi_0 (\bm{a}_t| \bm{s}_t)$ given by the prior trick to the observed rewards. This method was previously proposed under the name of \textit{entropy regularized + R} (ER) in \cite{czarnecki2019distilling}. However, the authors note that this approach suffers from high variance when using policy gradient-based algorithms.

Another approach is to add a regularization term $\ell_{\pi_0}(\theta, \bm{s}_t)$ to the policy gradient update rule, which is proportional to
\begin{equation}
    \label{update_rule}
    \mathbb{E}_{\pi, \mathcal{M}} \left[ \sum_{t=1}^T - \nabla_\theta \log \pi_\theta (\bm{a} | \bm{s}_t) \widehat{R}_t + \nabla_\theta \ell_{\pi_0}(\pi_\theta, \bm{s}_t) \right],
\end{equation}

where $\widehat{R}_t$ is an estimate of a baseline (usually the advantage function) and $\theta$ are the parameters of the policy. It can be shown \cite{czarnecki2019distilling} that we can recover a valid gradient field for the update rule above by adding a reward bonus of $-\ell_{\pi_0}(\pi_\theta, \bm{s}_{t+1})$. This is the idea behind their proposed \emph{expected entropy regularized + R} (E2R) algorithm, which sets $\ell_{\pi_0}(\pi_\theta, \bm{s}_t) = H^X_{\bm{s}_{t}} (\pi || \pi_0)$ and a reward bonus of $B^{E2R}(\bm{s}_t, \bm{a}_t) := \log \pi_0 (\bm{a}_{t+1}| \bm{s}_{t+1})$ - note that the cross entropy is the expectation of the bonus at the current step under the policy.

\section{Learning from Imperfect Priors}
\label{imperfect}

We turn our attention to the case in which the action prior $\pi_0$ is not optimal, but rather an imperfect prior providing helpful information within some parts of the state space. Traditional policy distillation techniques such as E2R cannot effectively leverage these types of priors to speed-up learning. We argue that the ability to take advantage of imperfect priors is essential because:
\begin{enumerate}
    \item providing useful guidance is often easier than providing expert action prior. For example, in a maze-solving scenario, it is much easier to produce an action prior that avoids running into a wall and visible dead-ends than a fully-capable maze solver;
    \item useful priors that humans seem to exploit, such as \emph{affordances} \cite{dubey2018investigating}, are not perfect. An action prior that informs affordances, i.e. which actions are more useful in each state, does not necessarily achieve high return when run as policy;
    \item finally, one is rarely, if not never, able to supply an expert prior over the entire state space.
\end{enumerate}
More precisely, we assume that the value 
of the prior $V^{\pi_0}$ 
may be small in parts of the state space $\mathcal{S}$. However, we also suppose that the prior contains useful information - at least partially - which can be leveraged to speed-up training. 

Typically, authors study distilling teachers with low returns by adding noise to their teachers. The commonly used approach - which we call \emph{random degradation} - is to independently sample a Bernoulli variable at each step that decides whether to use the expert or some noisy alternative. We argue that this way of degrading experts is not realistic because imperfect priors are usually tied to providing incorrect information in part of the state space rather than due to a noisy communication of that information.

In order to study more realistic imperfect priors, we make a second assumption that the prior's imperfections are tied to the state. Our proposed approach called \emph{structural degradation} selects a subset of the state space and replaces the teacher by some noisy alternative on those states. We expect priors that satisfy these two conditions to be more challenging to exploit than those who only meet the first.

In order to illustrate the challenges of learning from imperfect priors, suppose we wish to leverage a policy trained to open doors when training a more general agent that explores environments. Using policy distillation naively, the agent is heavily biased towards copying the prior, which may not produce desirable behaviour on the transfer task. A possible solution would be to add the reward bonus only when near a door, but it would require some heuristic criterion to decide when the bonus should be applied. We believe that a better solution would be to scale the bonuses according to their relevance to the task at hand, which is the approach we explore with our proposed algorithms.

\subsection{Adaptive Reward Bonuses}

In order to enable more robust transfer of action priors, we introduce \textit{prior weights} $\omega_{\pi_0}: \mathcal{S} \mapsto [0, 1]$ to scale the bonus at each state. These can be applied to any policy distillation algorithm to derive its adaptive variant. For example, the reward bonus and regularization loss for the Adaptive E2R (AE2R) algorithm is:
\begin{gather*}
    B_t^{AE2R} = \omega_{\pi_0} (\bm{s}_{t+1}) \log \pi_0 (\bm{a}_{t+1}, \bm{s}_{t+1}), \\
    \ell^{AE2R}_{\pi_0}(\pi, \bm{s}_t) = \omega_{\pi_0} (\bm{s}_{t}) H^X_{\bm{s}_t} (\pi || \pi_0).
\end{gather*}
Similarly, the adaptive variants of ER and E2R (named AER and AE2R respectively) can be derived. The weights are adjusted by minimizing the error between state value estimates and the ground-truth rewards. As a result, the method requires value function estimation. We believe that such limitation is not restrictive since most policy gradient methods already estimate state values for advantage estimation. Such state values are estimated with the help of a critic, which is trained by minimizing the following loss:
\begin{equation}
    \label{loss_v}
    \begin{split}
        &\mathcal{L}_V (\phi) = \\&\quad\mathbb{E}_{\bm{s}_t \sim \mathcal{B}} \left[ V_\phi(\bm{s}_t) - (R(\bm{s}_t, \bm{a}_t) + B_t + \gamma V_\phi(\bm{s}_{t+1})) \right]^2,
    \end{split}
\end{equation}
where $\phi$ are the critic's parameters and $\mathcal{B}$ is the batch distribution (i.e. the online distribution for on-policy methods). Note that the prior weights are included into $B_t$. 

We parameterize the prior weights $\omega_{\pi_0} (.)$ with parameters $\psi$ similarly to the policy and critics (e.g. with a neural network). The weights are learned by minimizing the same loss as the critic $\mathcal{L}_V$ with respect to $\psi$ instead, which we denote $\mathcal{L}_\omega (\psi)$. Such update rule procedure can be interpreted as learning to scale $\omega$ with the TD error between the critic and the value of the policy on the original MDP (without reward bonuses). Intuitively, the weight is adjusted according to how well it can explain the TD error of the policy on original task.
In the particular case of AE2R, the full prior weight loss is:
\begin{equation}
    \label{loss_a}
    \begin{split}
        &\mathcal{L}^{AE2R}_\omega (\psi) = \\&\quad\mathbb{E}_{\bm{s}_t \sim \mathcal{B}} \left[ E_t(\phi) - \omega_\psi(\bm{s}_{t+1}) \log \pi_0 (\bm{a}_{t+1}, \bm{s}_{t+1}) \right]^2,
    \end{split}
\end{equation}
where $E_t(\phi) := V_\phi(\bm{s}_t) - R(\bm{s}_t, \bm{a}_t) - \gamma V_\phi(\bm{s}_{t+1})$. The pseudocode for the E2R algorithm can be found on Appendix 1. 
Even though both the auxiliary loss $\ell_{\pi_0}$ and reward bonuses $B_t$ change throughout training, adaptive methods are also guaranteed to converge. Throughout an episode, the weight updates depend on the trajectory so far $\tau_t = (s_0, a_0, \dots, s_{t}, a_{t})$ which means that the auxiliary loss can be written as $\ell_{\pi_0} (\theta, \tau_t)$,. It can be shown that for all losses of this type, one can add a reward bonus of $B_t = - \ell_{\pi_0} (\theta, \tau_t)$ to recover a valid gradient field and, consequently, convergence guarantees. This result can be derived trivially from the proof of Theorem 2 in \citep{czarnecki2019distilling}.

\section{Related Work}

This work builds upon the Reinforcement Learning as Inference Framework \cite{levine2018reinforcement} in order to introduce priors into RL algorithms. To our knowledge, this paper is the first attempt at exploiting the action priors that naturally appear in the formulation of RL as inference. However, other works in the RL literature studied similar problem formulations aiming at improving sample efficiency by introducing domain-specific knowledge. These works can be broadly organized into four categories:

\textbf{Transfer Learning in Reinforcement Learning.} Given that our action priors can be interpreted as policy, it is natural to look for algorithms that learn from a teacher policy. This is precisely the problem studied in Transfer Learning in Reinforcement Learning \cite{lazaric2012transfer,zhu2020transfer}, and our method can be interpreted both as reward shaping \cite{ng1999policy} and policy distillation \cite{czarnecki2019distilling}. The naive reward bonus obtained via the RL as Inference approach was previously used in an approach called \emph{entropy regularized policy distillation} \cite{distral,SchulmanAC17}. However, our work introduces a novel estimator and adaptive approach in conjunction with the proposed reward bonus. 

\textbf{Meta-learning.} One of our goals is to use domain-specific knowledge to speed-up learning new tasks. Meta-RL \cite{duan2016rl,wang2016learning,finn2017model} leverages experience from previously seen tasks to achieve that same goal. In that sense, meta-learning can be interpreted as constructing a prior that is generally useful on a collection of tasks. While this framework has successfully achieved that goal, it is a compute-intensive approach that requires engineering tasks that teach the desired priors. It also usually encodes learned knowledge into the weights of neural networks, which limits its interpretability and usefulness.

\textbf{Learning from demonstrations.} Another way to speed-up reinforcement learning is providing (partial) trajectories that demonstrate desired behaviour \cite{schaal1997learning,vecerik2017leveraging,hester2018deep,nair2018overcoming,silver2018residual}. In terms of the framework developed in \cite{lazaric2012transfer}, learning from demonstrations is an \emph{instance-based transfer} while policy distillation is a \emph{representation transfer} - these are two ways to encode knowledge that have different costs and benefits. In cases where the desired priors are not already represented as trajectories, these methods introduce the challenge of inferring the prior information from the demonstrations. On the other hand, policy distillation requires expert teachers which may be a challenge to obtain.

\textbf{Action Advice in Multi-Agent Reinforcement Learning.} Since one can consider the soft action prior as a teacher, the proposed setting can be reinterpreted as collaborative learning in which a teacher advises a student \cite{da2019survey}. In this setting, a teacher may suggest an action to the student at each time step, and the student either chooses to take the teacher's advice or not. Some prior work in this setting studies learning when to \textit{ask for advice} and \textit{give advice} \cite{fachantidis2017learning, omidshafiei2019learning}, which is similar to our prior strength approach. These approaches often study providing advice under a constrained communication budget, differently from the soft action prior setting. Furthermore, these methods incorporate teacher's advice by following a teacher-driven control policy which is empirically known to degrade performance in policy distillation \cite{czarnecki2019distilling}.

\section{Experiments and Analysis}
\label{experiments}

We evaluate the performance of policy distillation methods when learning from teachers with imperfections. To this end, we study two experimental setups: the tabular GridWorld setting from \cite{czarnecki2019distilling} and the continuous control benchmarks from MuJoCo \cite{todorov2012mujoco} using OpenAI Gym \cite{openai_gym}.

The area ratio metric $r_{\pi_0}$ is employed to measure learning speed-up quantitatively. Inspired by other works in Transfer Learning in RL \cite{lazaric2012transfer}, we measure the area under the curve of returns obtained during evaluation episodes as a metric for learning speed. Let $A$ be the area under the curve when no prior information is given and $A_{\pi_0}$ be the area under the curve when distilling $\pi_0$ with some algorithm. Then, the area ratio is defined as $r_{\pi_0} = \frac{A_{\pi_0} - A}{A}$. In our reports, we normalize the area by the total number of timesteps so that the metric is comparable across experiments with different training budgets.

\subsection{Tabular Experiments on Grid Worlds}
\begin{figure*}[!t]
    \centering
    \includegraphics[width=\textwidth]{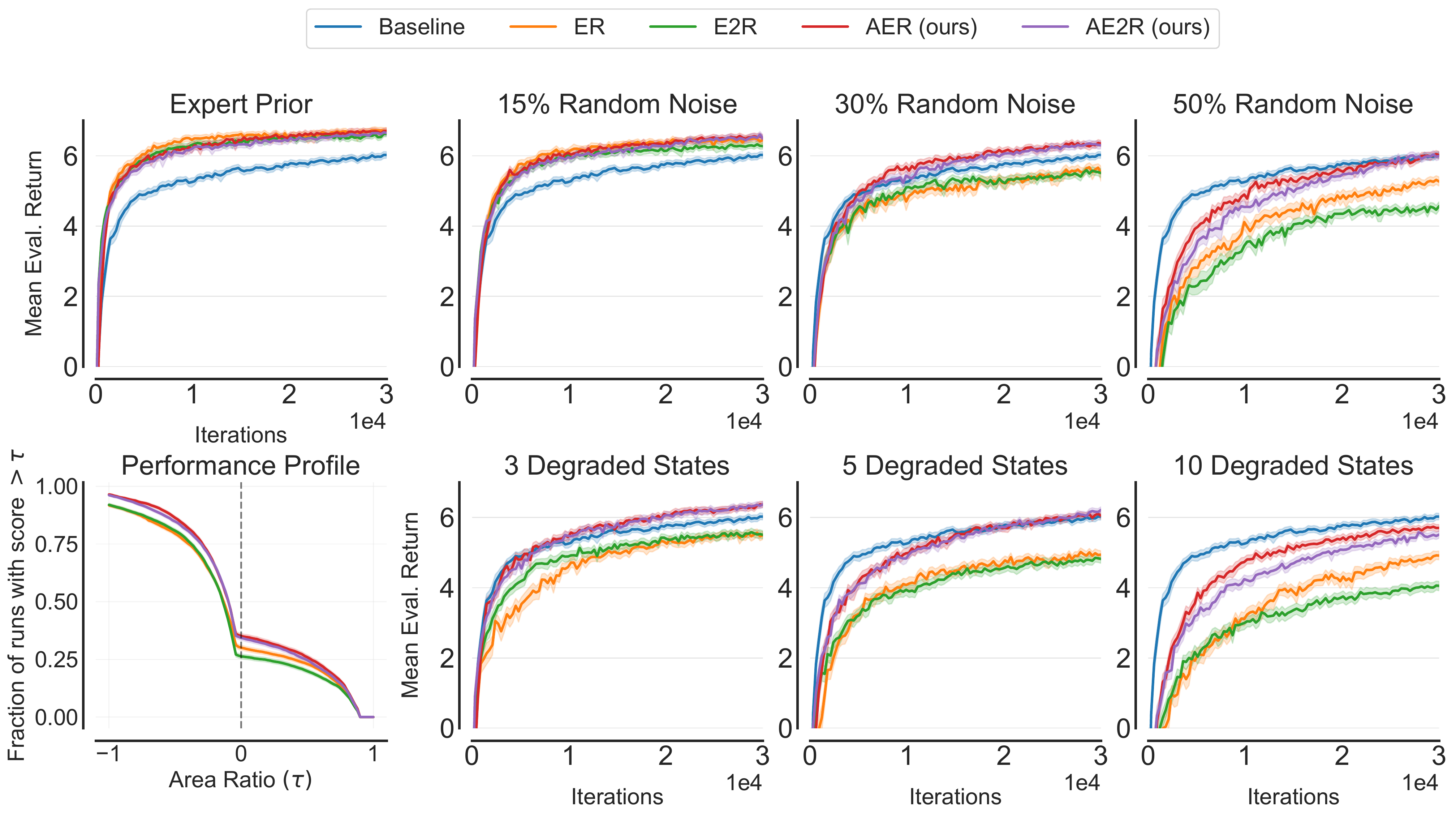}
    \caption{\textbf{Effect of prior degradations} Learning curves for all prior settings, obtained by averaging over one thousand grid worlds.  The leftmost column shows the expert prior setting (top) and performance profiles aggregating runs from all settings (bottom). The remaining plots show results for the random degradation (top) and structured degradation (bottom) settings, in which the quality of the prior decreases from left to right. In the performance profile, the dashed vertical line marks an area ratio of $0$, that is, no improvement over the baseline. Results show that adaptive methods are more robust to noisy priors but perform equally well as the baselines when learning from experts. Shaded bands represent pointwise $95\%$ confidence intervals. In all plots, the baseline is an actor-critic which does not leverage the prior.
    }
    \label{fig:tabular_results}
\end{figure*}

We closely follow the problem setting, MDP sampling procedure, and implementation details introduced in \cite{czarnecki2019distilling}. The environments consist of 20 x 20 Grid Worlds with walls which cannot be traversed and goal states with different associated rewards that may be terminal - additional details can be found in Appendix 2. Our motivation for this choice was two-fold:
\begin{itemize}
    \item \textbf{Ease of comparison} with other policy distillation methods from the literature;
    \item \textbf{Statistical relevance} of results. Due to computational limitations, deep RL experimentation would be limited to only a few runs, while this setup allows for averaging results over one thousand different MDPs.
\end{itemize}

Our experimental procedure is as follows: we associate each training run to a random seed which is used both for sampling the MDP and training. First, we sample a Grid World and train a Q-Learning agent with a budget of 30 thousand transitions. Then, we add noise to the Q-Learning policies to obtain our priors (except for the expert prior setting). Finally, we train each distillation algorithm with the different priors on all the sampled MDPs. On each run, we perform one hundred evaluation episodes throughout training to obtain the evaluation return curves, which are used for comparison between methods.

We use a basic actor-critic algorithm, in which we sample an episode according to the policy and update the policy parameters with the update rule in Eq. \ref{update_rule} with no auxiliary loss and a TD(1) estimate of the advantage for $\widehat{R}_t$. We chose this algorithm instead of Q-Learning as the baseline because it more closely resembles the actor-critic policy distillation algorithms. Policy distillation algorithms are implemented by adding reward bonuses and additional regularization losses to the actor-critic baseline. In all training curves, the baseline represents the base actor-critic which does not leverage priors.

Expert priors are degraded by swapping their action distributions by an \emph{adversarial policy} when a particular condition is met. The adversarial policy at a given state is obtained by simply using the negative Q-value of the expert to compute the action probabilities.

\paragraph{Expert setting} In the transfer learning literature, this setting corresponds to teacher distillation when using feedback for the environment. 
We compare the performance of the algorithms when learning from expert teachers in Figure \ref{fig:tabular_results}, which shows aggregate results for one thousand grid worlds. Results show that all methods perform similarly. As expected, there is no significant advantage to prior weights when learning from an expert. On the other hand, we verify that the proposed approach does not sacrifice performance for robustness in the expert setting.

\paragraph{Random degradation setting} In this setup, we randomly degrade the expert teachers from the expert setting. At each step, the degraded prior randomly selects either the expert or adversarial policy to return with a fixed probability which we call the \emph{prior noise}. A prior with $50\%$ random noise chooses between the expert and adversarial policies with equal probability. The learning curves for experiments with $15\%$, $30\%$, and $50\%$ prior noise are summarized in Figure \ref{fig:tabular_results}. It is clear that adaptive methods significantly outperform existing algorithms. Both ER and E2R are consistently outperformed by their adaptive counterparts.

\paragraph{Structural degradation setting} In this setting, the action prior is an expert prior that is degraded in some parts of the state space. That is, the prior is replaced by the adversarial policy (as defined previously) on a set of states. This setting is motivated by the discussion about realistic imperfect priors. In the random degradation setting, the agent might receive both good and bad advice while in the same state. In contrast, structural degradation guarantees that expert advice for degraded states is never seen throughout training. In order to choose which states to degrade, we employ the following procedure:
\begin{enumerate}
    \item gathering the value of each state according to the expert and using the \texttt{softmax} function to transform them into a probability distribution over states;
    \item sampling states from that probability distribution repeatedly until the desired number of unique states have been sampled.
\end{enumerate}
Using this sampling strategy ensures that the sampled states are relevant, that is, they will be encountered multiple times during training with high probability. Empirically, we observe that this procedure samples states close to the optimal path from the initial state to a rewarding terminal state with high probability. In order to better grasp the effect of degrading a certain number of states, we collected statistics about the training of experts: they encounter, on average, roughly 130 distinct states during training and produce trajectories with an average length of 10 steps during evaluation. 

The results from this experimental setup, presented in Figure \ref{fig:tabular_results}, highlight how the performance of policy distillation is brittle to structural noise in just a few states. The results show that introducing priors with as few as three degraded states is harmful when using existing policy distillation approaches, while adaptive algorithms still outperform the baseline. When using 5 and 10 degraded states, we observe that our adaptive methods are significantly more robust to the adversarial states than other methods.
\begin{table}[ht]
\centering
\caption{\textbf{Prior Weights in the Structural Degradation Setting}. Mean prior weights of degraded and non-degraded states encountered during evaluation episodes. We only present values for AE2R, since both AER and AE2R produce qualitatively similar results. The values after $\pm$ represent 95\% confidence intervals, computed as $1.96$ times the standard error of the mean.}
\label{tab:weights}
\resizebox{\linewidth}{!}{%
\begin{tabular}{lrr}
\toprule
          & \multicolumn{2}{c}{Mean Prior Weights} \\
Setting   & Non-Degraded States  & Degraded States \\ \midrule
3 states  & $0.432 \pm 0.001$       & $0.298 \pm 0.001$  \\
5 states  & $0.439 \pm 0.001$       & $0.299 \pm 0.001$  \\
10 states & $0.461 \pm 0.001$       & $0.308 \pm 0.001$  \\ \bottomrule
\end{tabular}%
}
\end{table}

\paragraph{Degraded states have lower prior weight.} We track the mean of the prior weights $\omega$ of degraded and non-degraded states encountered during evaluation episodes. Results show that adaptive methods assign smaller weights to degraded states when learning from a structurally degraded priors. Furthermore, we observe that the gap between the two means increases with the number of degraded states. These results support our intuition that adaptive methods assign lower prior weight to degraded states. Finally, note that prior weight is close to its initial value in states with a low number of visits, which skews these results  $0.5$.

\paragraph{Summarizing tabular results.} Figure \ref{fig:deep_figure} summarizes quantitative results across all settings. Our analysis concludes that adaptive algorithms achieve state-of-the-art performance, improving upon existing baselines when leveraging noisy priors. In other words, we observe that the proposed methods are more robust to suboptimal teachers.

\subsection{Deep Reinforcement Learning Experiments on Continuous Control Tasks}

Our deep RL experiments aim to evaluate how adaptive methods perform when distilling realistic teachers compared to their non-adaptive counterparts. To this end, we evaluate deep actor-critic algorithms using different distillation algorithms on continuous control environments from OpenAI Gym \cite{openai_gym}. Specifically, we perform evaluations on five environments from the MuJoCo \cite{todorov2012mujoco} benchmark with different levels of complexity: Hopper-v3, Walker2d-v3, HalfCheetah-v3, Ant-v3, and Humanoid-v3. Due to the difficulty of these tasks, we expect the trained experts to be farther away from optimality than in the tabular case. By using these teachers as priors, we evaluate whether robust policy distillation is helpful in realistic settings.

Following recommendations from prior work \cite{andrychowicz2020matters, flet2021learning}, we choose a PPO \cite{schulman2017proximal} with AVEC critic as our base algorithm. We present implementation details and hyperparameter choices in Appendix 3. In this setting, we estimate the prior weights $\omega$ with a neural network identical to the one employed by the critic. The agent learns by optimizing the sum of losses:
$$\mathcal{L}(\theta, \psi, \phi) = \mathcal{L}_{\pi} (\theta) + \mathcal{L}_V (\phi) + C_\omega \mathcal{L}_\omega (\psi),$$

where $\mathcal{L}_\pi$ is a loss that yields gradients proportional to Eq. \ref{update_rule}, and $C_\omega$ is an additional hyperparameter that controls the magnitude of weight updates. This formulation is equivalent to minimizing different objectives with different learning rates. We perform a coarse hyperparameter search over $\{ 1.0, 0.3, 0.03, 0.003 \}$ and report results for all values.

\begin{figure*}[!t]
    \centering
    \includegraphics[width=\textwidth]{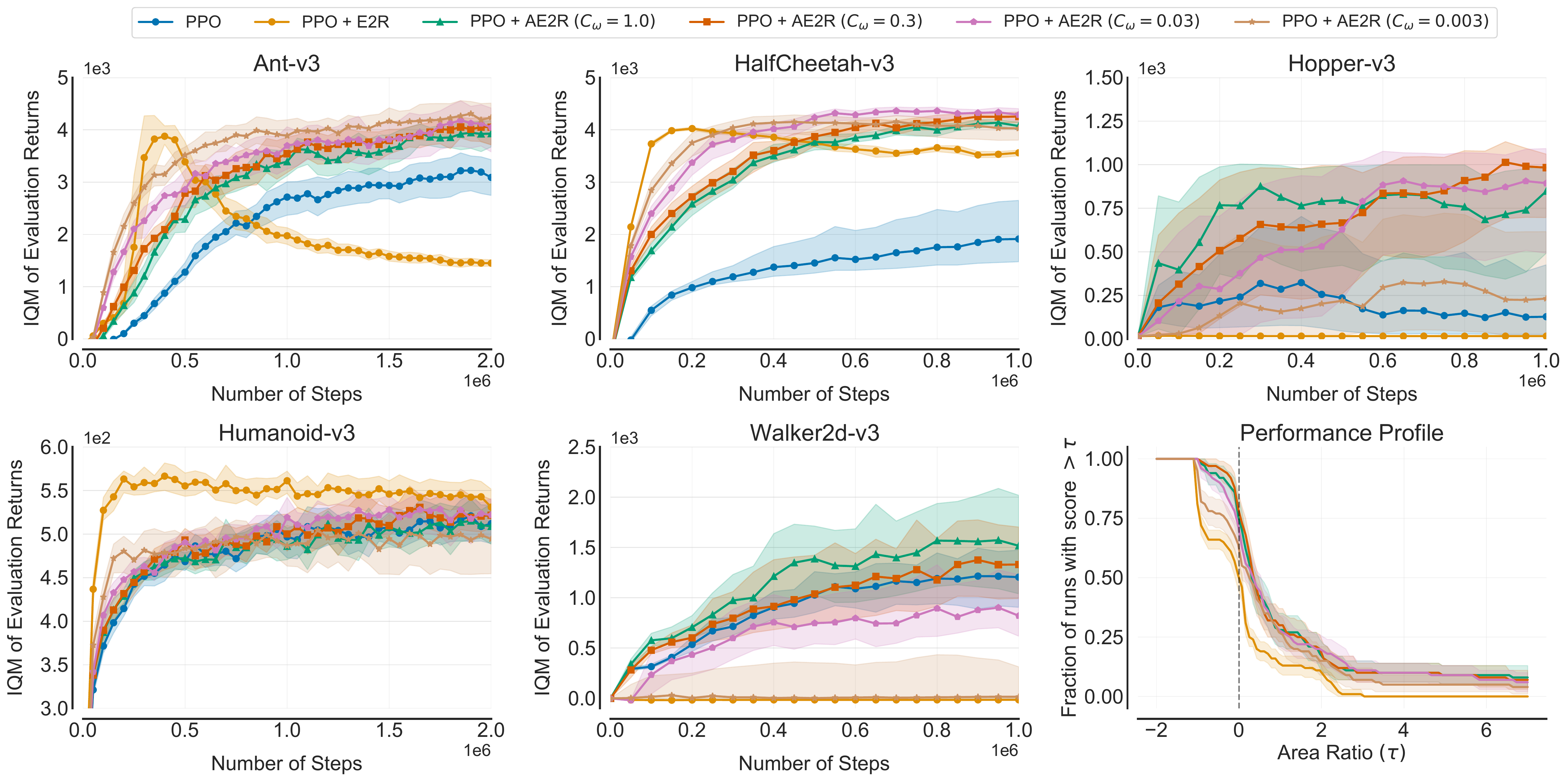}
    \caption{\textbf{Evaluation returns of PPO + E2R and PPO + AE2R (ours) compared to baseline PPO.} We report IQM of returns and performance profile of area ratios with 95\% stratified bootstrap CIs over 20 random seeds. In the performance profile, the dashed vertical line marks an area ratio of 0, that is, no improvement over the baseline. AE2R can consistently outperform the baseline, while E2R is unable to make progress in two environments and suffers from performance degradation. Furthermore, we observe that the PPO + AE2R approaches PPO + E2R as $C_\omega$ becomes smaller. Finally, the performance profile allows us to conclude that adaptive methods are able to speed-up learning with higher probability than standard distillation, even when using an expert as the teacher.}
    \label{fig:deep_figure}
\end{figure*}

Unlike the tabular case, we choose the best performing baseline over all seeds as the action prior. This selection criterion helps reduce the variance of distillation results which is essential since we perform fewer runs in this benchmark. We also follow recent recommendations \cite{agarwal2021deep} for reliable reporting of deep RL results in Figure \ref{fig:deep_figure}.

Note that we do not show results for PPO + ER and PPO + AER since these methods were unable to learn effectively in this setting, failing to improve upon a random policy. We believe that ER provides high-variance feedback in settings with high-dimensional continuous action spaces, whereas E2R is capable of mitigating this issue with the help from its auxiliary loss. We motivate our hypothesis by noting that increasing the variance of the bonus by artificially enlarging the state space has been shown to significantly hurt the performance of ER \cite{czarnecki2019distilling}.

Our findings suggest that current state-of-the-art policy distillation methods are brittle when used in conjunction with Deep RL in continuous control tasks. We observe that E2R tends to learn the teacher policy quickly and suffers performance degradation as training proceeds. Additionally, we note that as $C_\omega$ becomes smaller, AE2R approaches the results of E2R. As a result, lower values of $C_\omega$ generally work well in environments where E2R has strong performance. Conversely, we achieved the best results with $C_\omega = 1.0$ in the two environments in which E2R struggles to make progress.

By measuring the mean policy loss $\mathcal{L}_{\pi} (\theta)$ and auxiliary loss $\ell_{\pi_0}(\theta)$ of each algorithm throughout training, we find that PPO + E2R has losses of greater magnitude in most environments. Additionally, we find that policies trained with PPO + E2R tend to have significantly lower entropy than those obtained by our baseline PPO and PPO + AE2R. Given our findings, we believe that PPO + E2R excessively emphasizes feedback from the teacher. As a result, the performances obtained with PPO + E2R can vary significantly depending on the teacher policy and environment. A table containing the values of entropy and losses can be found in Appendix 4.

For example, we investigated the Humanoid-v3 environment and found that PPO + E2R performs well due to the scale of control costs in the ta. MuJoCo environments have negative rewards for penalizing large control forces. We believe that PPO + E2R incurs smaller penalties because it trains policies with considerable smaller entropy. In particular, the Humanoid-v3 environment has a large action space and high control costs by default, exacerbating this effect. We repeat our experiments on a modified Humanoid-v3 environment with 100 times smaller control costs. In Appendix 4, we show that the advantage of PPO + E2R disappears in the modified task.

Finally, Figure \ref{fig:deep_figure} shows the aggregate area ratio results across all environments. As expected, AE2R hurts performance (negative area ratio) significantly less often than E2R. Contrary to the tabular setting, the adaptive method outperforms its non-adaptive version when using an expert prior. This evidence supports our hypothesis that teachers are farther away from optimality in deep RL and motivates robust policy transfer algorithms for this setting.

\section{Discussion}

This paper explores the injection of non-expert action-related prior information into RL algorithms. Using the RL as Inference framework, we show that action priors can be generally introduced into MDPs via reward bonuses. Furthermore, the derived bonus corresponds to existing state-of-the-art policy distillation approaches.

Next, we discuss which types of imperfections are common in realistic priors and propose the structural degradation of priors to mimic these defects. We propose a novel class of adaptive algorithms to leverage realistic priors, designed to be robust to structural imperfections. The proposed algorithms weight the influence of the action prior on the learning process according to the prior's estimated usefulness at each state. 

Our tabular experimental study on one thousand sampled MDPs shows that the proposed adaptive algorithms are significantly more robust to suboptimal priors than existing state-of-the-art methods. We also perform a study of adaptive policy distillation with deep actor-critic algorithms on continuous control benchmarks, in which adaptive methods are more consistent while also outperforming the baseline in four of the five environments. More importantly, our work shows that accounting for imperfections when introducing priors into RL algorithms is essential in realistic scenarios.

Finally, this work opens doors to future efforts to leverage more general priors (such as affordances and hard-coded bots) to speed-up learning in RL algorithms. Exciting applications of these methods include improving sample efficiency in deep RL and lowering the computational requirements of RL benchmarks by providing background knowledge of the domain. 

\bibliography{aaai23.bib}

\clearpage

\appendix

\section{AE2R Pseudocode}
\label{pseudocode}

Below, we present the pseudocode for an Actor-Critic implementing AE2R for policy transfer. 

\begin{center}
\begin{algorithm}
    \DontPrintSemicolon
    \textbf{Input:} $\lambda_\pi \geq 0, \lambda_V \geq 0, \lambda_\omega \geq 0$ \;
    \textbf{Initialize:} parameters $\theta$, $\phi$ and $\psi$ \;
    \For{\textup{each update step}}{
        batch $\mathcal{B} \gets \emptyset$ \;
        \For{\textup{each environment step}}{
            $a_t \sim \pi_\theta (s_t)$ \;
            $s_{t+1} \sim p (s_t, a_t)$ \;
            $\mathcal{B} \gets \mathcal{B} \cup \{(s_t, a_t, r_t, s_{t+1})\}$ \;
        }
        \For{\textup{each gradient step}}{
            Update $\theta$ according to the update rule in Equation \ref{update_rule} with $B^{AE2R}$ and $\ell^{AE2R}$ \;
            $\phi \gets \phi - \lambda_V \hat{\nabla}_\phi \mathcal{L}_V^{AE2R}(\phi)$ (see Equation \ref{loss_v}) \;
            $\psi \gets \psi - \lambda_\omega \hat{\nabla}_\psi \mathcal{L}_\omega^{AE2R} (\psi)$ (see Equation \ref{loss_a}) \;
        }
    }
    \caption{Simple Actor-Critic implementation of AE2R.}
\end{algorithm}
\end{center}

Note that the prior weight parameters are updated \textit{after} the critic.

\section{Grid World Experimental Setup}
\label{exp_appendix}

\begin{figure}[h]
    \centering
    \includegraphics[width=0.7\linewidth]{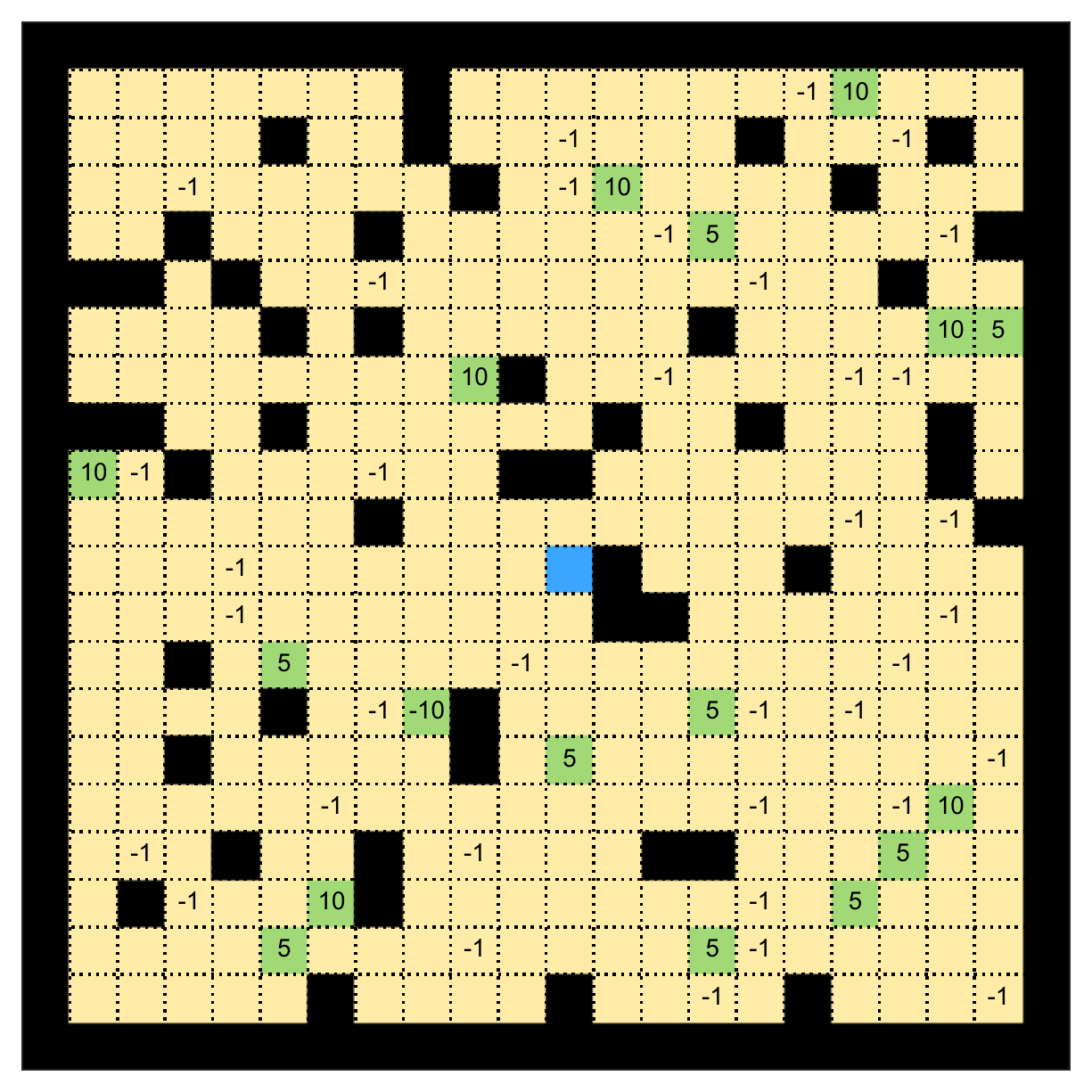}
    \caption{An example Grid World environment. The initial state, walls and terminal states are represented with blue, black and green tiles, respectively. Non-zero rewards are indicated on their corresponding state.}
    \label{fig:environment}
\end{figure}

As stated in the main text, we followed the experimental setup of \cite{czarnecki2019distilling} closely. This section provides additional details about our experimental setup for the Grid World tabular experiments:

\textbf{Environment.} We sample grid worlds (that is, MDPs) according to the sampling procedure specified in \cite{czarnecki2019distilling}. The procedure traverses a 20 x 20 grid and samples an object to place at each position. Objects may be walls, empty spaces, and objectives (which can give -10, -5, -1, 1, 5, or 10 as a reward). Objectives with -10, -5, +5, and +10 rewards are terminal, and the initial state is set to the middle of the grid. Figure \ref{fig:environment} shows a Grid World example. Furthermore, there is a probability of $0.01$ of termination at each step and a transition noise of $0.1$ per step, replacing the agent's action with a uniformly sampled one. Lastly, observations are represented as 9 x 9 grids of vision centered on the agent.

\textbf{Parameterization.} Both the Q-Learning baseline and the actor-critics employ tabular models; that is, they have distinct parameters for each observation. Policies and Q tables have four parameters per observation (one for each action), while critics only have one (for the state-value). All parameters are initialized to zero.

\textbf{Evaluation.} Curves are made with data from 100 evaluation episodes evenly spaced throughout training, that is, every 300 update steps. The evaluation data is collected with deterministic policies.

\textbf{Hyperparameters} All hyperparameters not mentioned in our work are set to their default value in the original setup. The supplementary code \footnote{which can be found at our accompanying repository at .} contains a complete implementation of the algorithms, environment and plotting scripts.

\textbf{Computing Infrastructure.} Tabular agents are trained on a 48-core system equipped with Intel Xeon E5-2687W v4 CPUs at 3.00Ghz. Each experiment (that is, one thousand runs) takes less than an hour to complete. Parallelization of training runs across processes in essential to achieve short training times.

When using the Q-Learning agent as an teacher, we compute log-probabilities with a Boltzmann policy with temperature $\tau$:

$$\pi_0(\bm{a} | \bm{s}) = \frac{\exp \left(Q(\bm{s}, \bm{a}) / \tau \right)}{\sum_{\bm{a}' \in \mathcal{A}} \exp \left(Q(\bm{s}, \bm{a}') / \tau \right)}.$$

In all experiments, we use a temperature of $\tau=1$ when calculating the policy from Q-values and TD(1) advantage estimates. We found that other choices yielded qualitatively similar results, so we restrict our experiments to these choices of temperature and advantage estimates for simplicity.


\section{Additional Visualizations of Tabular Results}

\begin{figure}[ht]
    \centering
    \includegraphics[width=\linewidth]{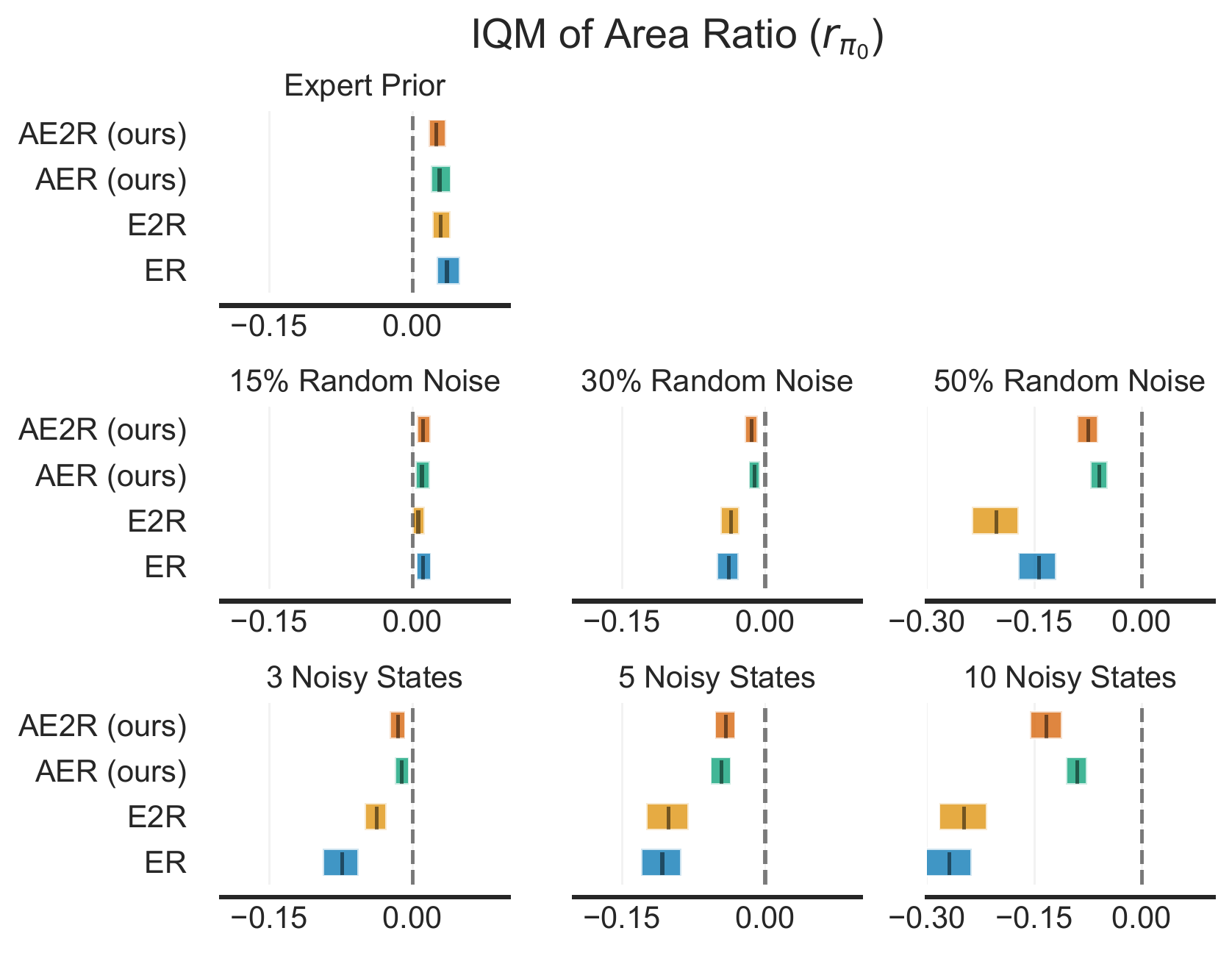}
    \caption{IQM of the area ratio $r_{\pi_0}$ per algorithm for each prior setting. Shaded bars represent bootstrapped 95\% CIs. }
    \label{fig:agg_metrics}
\end{figure}

We provide additional visualizations of results on the tabular setting. Specifically, we compare the aggregate area ratio $r_{\pi_0}$ across all seeds per prior setting. Table \ref{tab:area_ratio} shows the inter-quartile mean (IQM) with 95\% stratified bootstrap confidence intervals of the area ratio $r_{\pi_0}$ per algorithm for each setting. Finally, Figure \ref{fig:agg_metrics} summarizes the quantitative results.

\begin{table*}[t]
\centering
\caption{IQM of the area ratio $r_{\pi_0}$ (greater is better) on one thousand grid worlds for ER, AER, E2R and AE2R. The prior $\pi_0$ depends on the setting, which are: Expert Prior (EP), Random Degradation (RD), and Structural Degradation (SD). Regarding this metric, larger values are better and positive values (resp.\@ negative values) indicate learning faster (resp.\@ slower) than the baseline (which does not leverage prior information). Values in parenthesis indicate bootstrapped 95\% confidence intervals and the best performance is presented in boldface.}
\label{tab:area_ratio}
\resizebox{\linewidth}{!}{%
\begin{tabular}{ll|rrrr}
\toprule
Setting & \multicolumn{5}{c}{Area Ratio $r_{\pi_0}$ ($\times 10^{-3}$)} \\ 
& & ER & E2R & AER & AE2R \\ \midrule
EP & & $\mathbf{36 \ (25, 50)}$ & $29 \ (21, 40)$ & $28 \ (19, 40)$ & $25 (17, 35)$ \\ \midrule
 & 15\% & $\mathbf{11 \ (4, 19)}$ & $6 \ (0, 13)$ & $10 \ (3, 18)$ & $\mathbf{11 (4, 19)}$ \\
RD & 30\% & $-38 \ (-51, -28)$ & $-36 \ (-47, -27)$ & $\mathbf{-11 \ (-17, -5)}$ & $-14 \ (-21, -8)$  \\
 & 50\% & $-144 \ (-172, -119)$ & $-203 \ (-237, -172)$ & $\mathbf{-59 \ (-72, -48)}$ & $-74 \ (-91, -61)$ \\ \midrule
 & 3 & $-74 \ (-94, -57)$ & $-38 \ (-50, -27)$ & $\mathbf{-11 \ (-19, -4)}$ & $-16 \ (-24, -8)$ \\
SD & 5 & $-108 \ (-130, -88)$ & $-101 \ (-125, -80)$ & $-46 \ (-57, -36)$ & $\mathbf{-41 \ (-53, -31)}$ \\
 & 10 & $-269 \ (-303, -237)$ & $-248 \ (-283, -216)$ & $\mathbf{-90 \ (-106, -76)}$ & $-133 \ (-156, -112)$ \\ \bottomrule
\end{tabular}%
}
\end{table*}

\section{Continuous Control Experimental Setup}
\label{deep_appendix}

This section further details our experiments on continuous control tasks in the deep RL setting.

\textbf{Environments.} We use the MuJoCo \cite{todorov2012mujoco} continuous control environments on Open AI Gym \cite{openai_gym} as our benchmark. In particular, we take inspiration from recent benchmarks \cite{andrychowicz2020matters} and train agents on the Ant-v3, HalfCheetah-v3, Hopper-v3, Humanoid-v3, and Walker-v3 environments. Finally, we train agents from one million timesteps on all environments except Ant-v3 and Humanoid-v3, in which agents are trained for two million timesteps.

\textbf{Implementation.} We modify a base continuous PPO implementation \cite{huang2021cleanrl} in PyTorch \cite{pytorch} to support E2R and AE2R policy distillation.

\textbf{Model Architecture.} Policies are parameterized with diagonal Gaussian distributions over the continuous action space. The policy diagonal covariance matrix is a learnable parameter, not dependent on the state. We parameterize the policy mean, critic and prior weights with dense neural networks made of two hidden layers of size 64 and \texttt{tanh} activations. The size of the final layer is the same as the size of the action space for the policy and one for the critic and prior weights. The weights of the last layers are initialized with 100 times smaller scale for the policy and prior weight networks.

\textbf{Evaluation.} We evaluate agents every 50 thousand timesteps by freezing the policy and reporting the mean undiscounted reward of the stochastic policy on 50 evaluation episodes.

\textbf{Hyperparameters.} The hyperparameters used in our PPO implementation are presented in Table \ref{tab:hyperparams}. Data collected during a rollout is randomly assigned to minibatches at the start of each epoch.

\textbf{Computing Infrastructure.} Deep RL experiments were performed on a 32-core system equipped with Intel Xeon Gold 6134 CPUs at 3.20GHz and NVIDIA GeForce RTX 2080Ti GPUs. The training time varied from 30 minutes to 2 hours per run, depending on the environment and algorithm.

\begin{table}[t]
\centering
\caption{Hyperparameters for PPO used in the continuous control experiments. The values in parenthesis were used exclusively with the Hopper-v3 environment, which required extra tuning for consistent results within the training budget.}
\label{tab:hyperparams}
\resizebox{\linewidth}{!}{%
\begin{tabular}{lr}
    \toprule
    Hyperparameter & Setting \\ \midrule
    Nb. of epochs & 10 \\
    Clipping parameter ($\epsilon$) & 0.2 \\
    Learning rate & $3 \times 10^{-4}$ \\
    Optimizer & Adam \cite{kingma2014adam} \\
    Observation Normalization & Yes \\
    Advantage Normalization & Yes \\
    Discount factor & 0.99 \\
    GAE $\lambda$ & 0.95 \\
    Nb. of environments & 8 (2) \\
    Rollout size & 1024 (2048) \\
    Batch size & 128 (64) \\
    Value Loss & AVEC \cite{flet2021learning} \\
    \bottomrule
\end{tabular}%
}
\end{table}

\section{Additional Continuous Control Results}
\label{metrics_appendix}

\begin{table*}[ht]
\centering
\caption{\textbf{Metrics for Continuous Control Experiments.} In order to compare learning dynamics, we compute the mean of different training metrics throughout training. For ease of comparison, we present the difference between the metric during distillation and the metric for baseline PPO. Intuitively, the differences show the change caused by introducing a prior with a given algorithm. We observe that PPO + E2R has a larger impact on the losses and entropy of trained policies, which supports our hypothesis that non-adaptive method place greater emphasis on distilling the teacher.}
\label{tab:training_metrics}
\resizebox{\linewidth}{!}{%
\begin{tabular}{lrrrrll}
\toprule
                & \multicolumn{2}{c}{Policy Loss Difference} & \multicolumn{2}{c}{Entropy Difference} & \multicolumn{2}{c}{Auxiliary Loss} \\17
Environment     & PPO + E2R            & PPO + AE2R          & PPO + E2R          & PPO + AE2R        & PPO + E2R             & PPO + AE2R         \\ \midrule
Ant-v3          & $0.273 \pm 0.153$       & $0.006 \pm 0.023$      & $-22 \pm 15$          & $-1.0 \pm 11$        & $-1.7 \pm 62$         & $0.6 \pm 12$          \\
HalfCheeetah-v3 & $0.024 \pm 0.030$       & $0.009 \pm 0.027$      & $-13 \pm 6.4$         & $-2.6 \pm 3.9$       & $-0.2 \pm 4.9$        & $0.1 \pm 1.7$         \\
Hopper-v3       & $0.026 \pm 0.036$       & $0.009 \pm 0.024$      & $-6.7 \pm 1.9$        & $-2.3 \pm 1.2$       & $1.7 \pm 0.3$         & $0.10 \pm 0.08$       \\
Humanoid-v3     & $0.423 \pm 0.051$       & $0.014 \pm 0.029$      & $-21 \pm 20$          & $1.4 \pm 17$         & $459 \pm 1580$        & $18 \pm 520$          \\
Walker-v3       & $0.033 \pm 0.037$       & $0.004 \pm 0.029$      & $-14 \pm 5.3$        & $-2.2 \pm 3.1$       & $0.87 \pm 3.4$        & $0.076 \pm 1.2$        \\ \bottomrule
\end{tabular}%
}
\end{table*}

In this section, we compare the policy loss and entropy of PPO + E2R and PPO + AE2R throughout training. Table \ref{tab:training_metrics} shows the difference of each metrics mean to the PPO baseline, for each environment. Additionally, the auxiliary loss $\ell_{\pi_0}(\theta)$ is also reported. First, we observe that the policy loss difference is a couple of orders of magnitude greater for E2R than AE2R across most environments. Next, the mean of entropy differences is significantly more negative for the non-adaptive method, which indicates that policies learned with E2R are much less stochastic. Finally, the scale of the auxiliary losses $\ell_{\pi_0}(\theta)$ is greater for E2R than AE2R, which allows us to conclude that greater emphasis is given to copying the teacher - that is, minimizing $\ell_{\pi_0}(\theta)$ - when employing PPO + E2R.

\begin{figure}
    \centering
    \includegraphics[width=\linewidth]{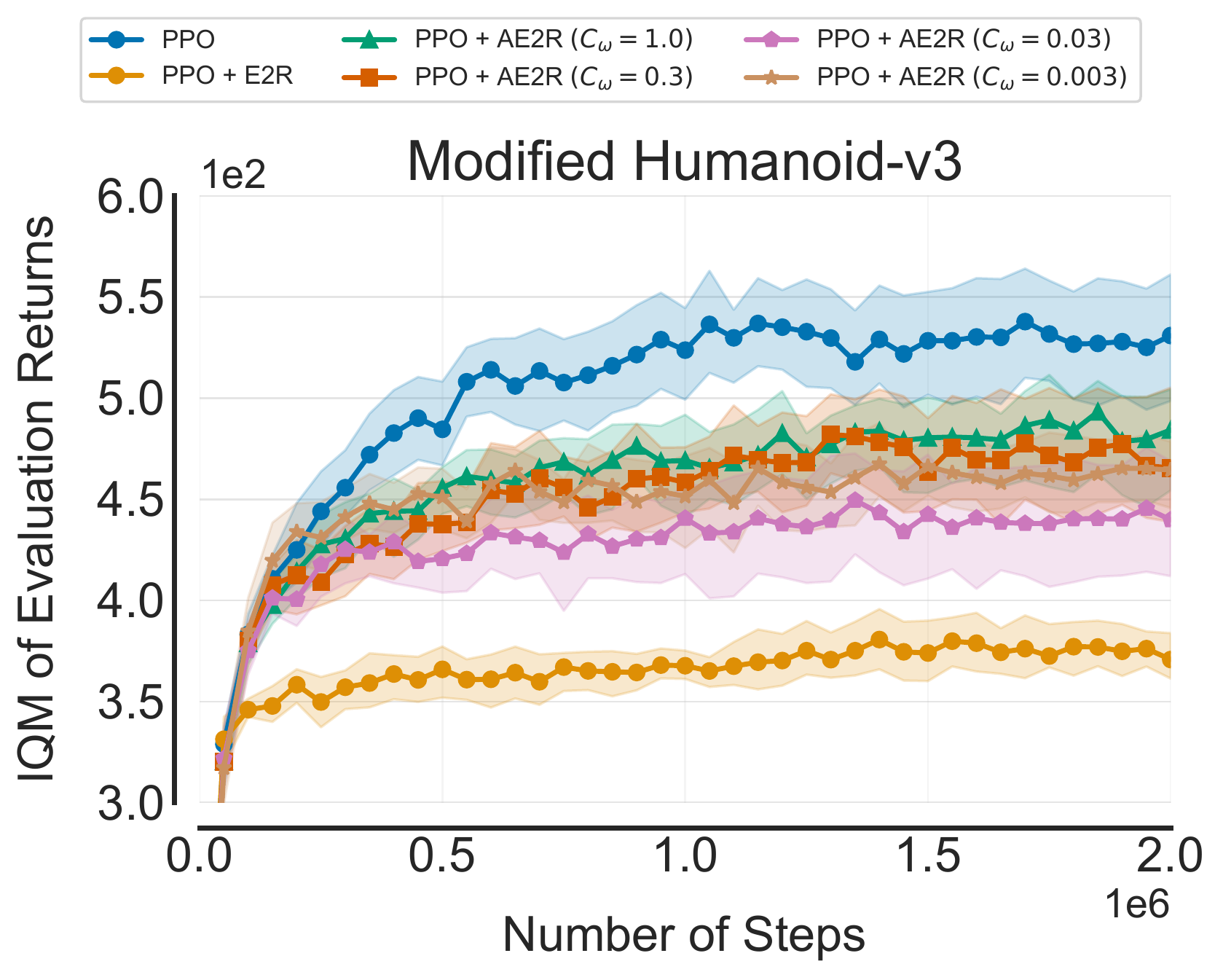}
    \caption{\textbf{Evaluation returns on the modified version of Humanoid-v3.} In this experiment setup, we set the coefficient that controls the scale of control cost penalties to a 100x smaller value than its default. Results show that PPO + E2R underperforms other methods, as opposed to the results obtained in the original environment. These results illustrate the brittleness of PPO + E2R.}
    \label{fig:appendix_humanoid}
\end{figure}

As discussed in the main text, we found that the emphasis on distilling the teacher seemed to help PPO + E2R achieve good performance in tasks with high control costs. In order to test this hypoothesis, we repeated Deep RL experiments on a modified Humanoid-v3 environment with 100x smaller control costs. Figure \ref{fig:appendix_humanoid} shows evaluation return curves for PPO, PPO + E2R and PPO + AE2R on this new task. Results support our hypothesis, as PPO + E2R underperforms every other method in the modified setup. These findings suggest that adaptive scaling of teacher feedback is also important for robustness across tasks as well.

\end{document}